\documentclass[10pt,twocolumn,letterpaper]{article}

\usepackage{iccv}              

\definecolor{iccvblue}{rgb}{0.21,0.49,0.74}
\usepackage[pagebackref,breaklinks,colorlinks,allcolors=iccvblue]{hyperref}
\usepackage{multirow}
\definecolor{dg}{HTML}{32CD32}

\def\paperID{6007} 
\def\confName{ICCV}
\def\confYear{2025}

 \title{OrderChain: Towards General Instruct-Tuning for Stimulating \\ the Ordinal Understanding Ability of MLLM}
\author{Jinhong Wang\textsuperscript{\rm 1,2,3$*$} \quad Shuo Tong\textsuperscript{\rm 2$*$} \quad Jian Liu\textsuperscript{\rm 3} \quad Dongqi Tang\textsuperscript{\rm 3} \quad Weiqiang Wang\textsuperscript{\rm 3} \quad Wentong Li\textsuperscript{\rm 1} \\ Hongxia Xu\textsuperscript{\rm 2} \quad Danny Z. Chen\textsuperscript{\rm 4} \quad Jintai Chen\textsuperscript{\rm 5$\dagger$} \quad Jian Wu\textsuperscript{\rm 2$\dagger$} \vspace{0.3cm} \\
    \textsuperscript{\rm 1}College of Computer Science \& Technology, Zhejiang University\\
    \textsuperscript{\rm 2}Transvascular Implantation Devices Research Institute and Liangzhu Laboratory \\
        \textsuperscript{\rm 3}Ant Group
    \quad \quad \textsuperscript{\rm 4}University of Notre Dame
    \quad \quad \textsuperscript{\rm 5}HKUST (Guangzhou) \\  
{\small{{$^{*}$}Contribute equally~~{$^{\dag}$}Corresponding Authors}}
}

\begin{document}
\maketitle
\begin{abstract}
Despite the remarkable progress of multimodal large language models (MLLMs), they continue to face challenges in achieving competitive performance on ordinal regression (OR; \textit{a.k.a.} ordinal classification). To address this issue, this paper presents OrderChain, a novel and general prompting paradigm that improves the ordinal understanding ability of MLLMs by \texttt{specificity} and \texttt{commonality} modeling. Specifically, our OrderChain consists of a set of task-aware prompts to facilitate the specificity modeling of diverse OR tasks and a new range optimization Chain-of-Thought (RO-CoT), which learns a commonality way of thinking about OR tasks by uniformly decomposing them into multiple small-range optimization subtasks. Further, we propose a category recursive division (CRD) method to generate instruction candidate category prompts to support RO-CoT automatic optimization. Comprehensive experiments show that LLaVA model with our OrderChain improves baseline LLaVA significantly on diverse OR datasets, e.g., from {\bf \textit{47.5\%}} to {\bf \textit{93.2\%}} accuracy on the Adience dataset for age estimation, and from {\bf \textit{30.0\%}} to {\bf \textit{85.7\%}} accuracy on the Diabetic Retinopathy dataset. Notably, LLaVA with our OrderChain also remarkably outperforms state-of-the-art methods by {\bf \textit{27\%}} on accuracy and  {\bf \textit{0.24}} on MAE on the Adience dataset. To our best knowledge, our OrderChain is the first work that augments MLLMs for OR tasks, and the effectiveness is witnessed across a spectrum of OR datasets. Project Page: \url{https://order-chain.github.io/}.
\end{abstract}    
\section{Introduction}
\label{sec:intro}

\begin{figure}[t]
\centering
\includegraphics[width=0.48\textwidth]{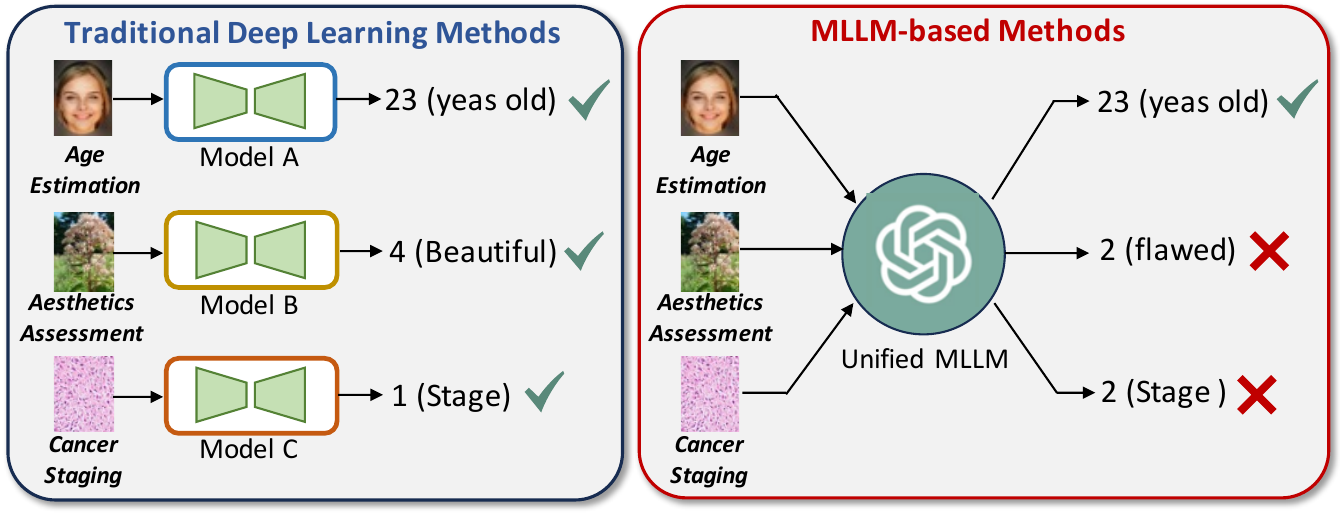}
\caption{Comparing traditional ordinal regression (OR) methods and modern MLLM methods.  Traditional OR methods perform well but need to train separate models for different tasks. MLLMs can be unified for diverse tasks but struggle in performance. }
\label{fig1}
\vskip -1 em
\end{figure}

With the success of Large language models (LLMs), e.g., GPT3~\cite{brown2020language}, LLaMA \cite{touvron2023llama}, Gemini~\cite{team2023gemini}, and Qwen~\cite{bai2023qwen}, researchers have been enhancing these models by incorporating visual understanding capabilities and led to emergence of multimodal large language models (MLLMs), such as LLaVA~\cite{liu2024improved} and GPT-4V~\cite{gpt4v} and Qwen-VL~\cite{bai2023qwen,bai2023qwen-vl}, exhibiting capabilities comparable to human intelligence in visual understanding tasks~\cite{yuan2023osprey,wu2024dettoolchain}. Despite these advances, the potential of MLLMs on order understanding for ordinal regression tasks is not yet well explored.

Ordinal regression (OR) refers to classifying object instances into ordinal categories, and is crucial for applications in various areas like facial age estimation~\cite{levi2015age,liu2018constrained,shin2022moving}, image aesthetics assessment~\cite{kong2016photo, lee2019image, pan2019image}, medical disease grading~\cite{liu2018ordinal,gentry2015penalized} and so on. The category labels of these tasks all follow a natural order. Unlike general classification tasks, ordinal understanding is a crucial issue for the representation learning of OR tasks. Mainstream methods in the past, including order distribution learning~\cite{liu2019probabilistic,li2021learning}, instance comparison~\cite{shin2022moving}, and CLIP-based~\cite{li2022ordinalclip,wang2023learning,du2024teach}, all revolve around this point. Although these methods are effective, it is still challenging to train a unified model for all OR tasks due to different specificities of diverse tasks (e.g., in terms of the number and range of categories). Therefore, separate models are still needed for different tasks (see Fig.~\ref{fig1}).

\begin{figure}[t]
\centering
\includegraphics[width=0.49\textwidth]{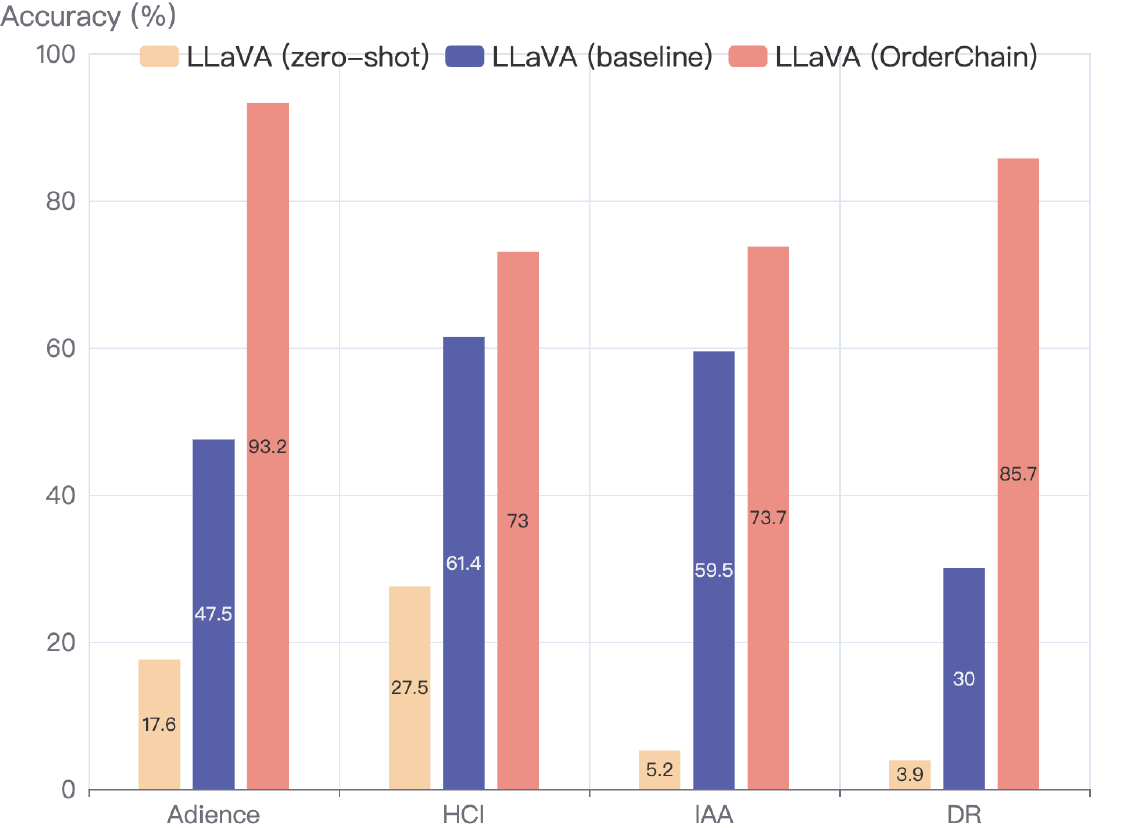}
\caption{The performance of different versions of LLaVA on diverse datasets.  The accuracy of Vanilla LLaVA is less than 60\% on most datasets. Adience: Age estimation dataset. HCI: Historical color image dating dataset. IAA: Image aesthetics assessment dataset. DR: Diabetic retinopathy grading dataset.}
\label{fig22}
\vskip -1 em
\end{figure}

MLLMs appear promising to address this challenge with their extensible language system. \textcolor{black}{However, through investigation, we find that no matter whether MLLMs are training-free (zero-shot) for direct inference or are fine-tuned by image-label pairs (baseline), their performance is not satisfactory, as shown in Fig.~\ref{fig22}.} The main challenges that hinder the model performance can be attributed to: (1) Lack of specificity modeling. Especially, when using MLLM for zero-shot inference, the conceptual information of domain knowledge and prior knowledge of category boundaries and range are overlooked, which contain task-related hints for task-aware ordinal understanding. (2) Lack of commonality modeling. For OR tasks, their most striking similarity is that the categories have an order. Thus, it is crucial to learn a commonality way of thinking about OR tasks which can strengthen MLLMs' understanding of category order.

\begin{figure*}[t]
\centering
\includegraphics[width=0.95\textwidth]{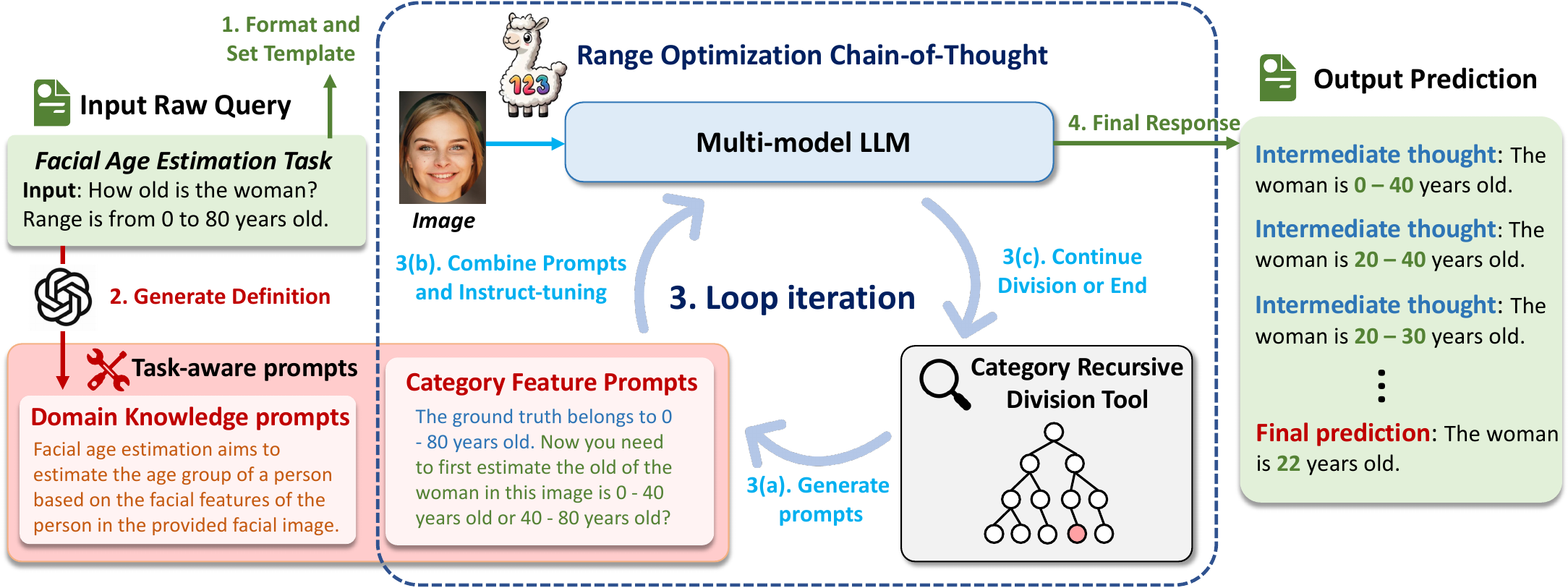}
\caption{The overall framework of our proposed OrderChain for improving the ordinal understanding of Multi-modal LLMs (e.g., LLaVA). }
\label{fig2}
\end{figure*}

To tackle these two challenges, this work investigates how the ordinal understanding capability of MLLMs can be realized by developing a novel Chain-of-Thought method, called OrderChain. OrderChain is driven by three pivotal motivations. {\bf First}, domain-related knowledge needs to be utilized to guide the MLLM to extract critical imaging features and make predictions more effectively. {\bf Second}, for different OR tasks, the knowledge on the number and range of candidate category labels needs to be injected to mitigate the out-of-bound predictions made by the MLLM. Both these two types of knowledge can help the MLLM to model the specificity of different OR tasks. {\bf Third}, inspired by the idea of decomposition, OR tasks can be decomposed into sub-interval classification tasks. Through continuous decomposition and classification, the candidate range can be gradually optimized to obtain the final prediction. This insight is applicable to all tasks whose categories have an order, which can help the MLLM to model the commonality of OR tasks. Based on these considerations, our proposed OrderChain consists of the following key components: {\bf (1) Customized \textit{task-aware} prompts to guide MLLMs in specificity modeling.} These include domain knowledge prompts, which provide prior knowledge to help understand the task, and category feature prompts, which offer information on the number and range of candidate labels to enable more accurate predictions. {\bf (2) A Range Optimization Chain-of-Thought (RO-CoT) for progressive ordinal regression.} As shown in Fig.~\ref{fig2}, RO-CoT transforms inputs into instruction templates, generates task definitions as domain knowledge prompts, recursively divides labels or coarse predictions into smaller subsets as category feature prompts, and iteratively refines predictions until a final response is achieved. This approach decomposes visual rating tasks into subtasks for gradual refinement. {\bf (3) A \textit{category recursive division} method for coarse-to-fine candidate label generation.} This method automatically splits RO-CoT predictions into refined subsets, which are used as category feature prompts to support continuous optimization in the prediction process.

Compared to traditional deep learning (DL) methods, Our OrderChain allows MLLMs to train a unified model for all OR tasks. Extensive experiments on OR datasets of various domains show the effectiveness of our OrderChain and its components. Notably, OrderChain achieves state-of-the-art performance in facial age estimation tasks, remarkably improving accuracy to 93.2\% (a 27\% improvement). In other domains of OR tasks, LLaVA with our OrderChain yields improvement of $\sim$12\% to $\sim$56\% compared to baseline LLaVA, showing highly competitive performance.

Our main contributions are summarized as follows.
\begin{enumerate}[{\bf (A)}]
\item For the first time, we explore the potential of MLLMs for ordinal regression tasks.

\noindent
    \item We propose a new prompting paradigm, called OrderChain, to instruct the MLLM to manage OR tasks in a general and progressive manner with a range optimization Chain-of-Thought.
    
\noindent
    \item We design a set of task-aware prompts, including domain knowledge prompts and category feature prompts for enhancing MLLM task-specific knowledge modeling. 
    
\noindent
    \item Extensive experiments show the effectiveness of our OrderChain on various OR datasets, which provides a promising way to establish a unified OR model.
\end{enumerate}

\section{Related Work}
\label{sec:formatting}

\subsection{Multimodal Large Language Models (MLLMs)} 

Large language models (LLMs), such as GPT-3~\cite{openai2023gpt4}, Qwen~\cite{qwen}, 
and LLaMA~\cite{touvron2023llama}, have attracted lots of attention for their remarkable capabilities across various linguistic tasks. This wave of interest has paved the way for the development of recent Multimodal Large Language Models (MLLMs), which integrate LLMs with visual encoders to enable an enriched comprehension and understanding of multimodal content. Prominent examples include the LLaVA series~\cite{liu2023llava,llava1.5,llavanext},
GPT-4V~\cite{gpt4v}, mPLUG-Owl~\cite{ye2023mplug},  InstructBLIP~\cite{instructblip}, Qwen-VL~\cite{bai2023qwen-vl}, Google’s Gemini series~\cite{team2023gemini, reid2024gemini1_5}, etc. Based on LoRA~\cite{hu2022lora}, MLLMs can be fine-tuned for downstream generalization in various areas, such as detection~\cite{wu2024dettoolchain} and segmentation~\cite{yuan2024osprey}. These advances highlight the diverse and expanding landscape of MLLMs, which have remarkably impacted the landscape of Artificial General Intelligence (AGI). This work is the first to explore the potential of MLLMs for OR tasks, providing a promising way to build a unified OR model.

\subsection{Chain-of-Thought (CoT)} 
CoT prompting is a specialized tool for inducing LLMs to produce intermediate reasoning steps that lead to a final answer and decision-making~\cite{wei2022chain}.
This technique elicits LLMs to generate a coherent series of intermediate reasoning steps that arrive at the final answer to a question. The traditional prompting method~\cite{brown2020language} performs poorly when it faces tasks that require reasoning abilities. Inspired by the concept of using intermediate steps to solve reasoning problems~\cite{cobbe2021training}, the chain-of-thought method mimics a step-by-step thinking process and breaks a multi-step problem into intermediate steps, enabling the model to deduce
more accurate results~\cite{wei2022chain}. Moreover, CoT is also a very effective tool for applying LLMs to a variety of downstream scenarios. Diverse customized CoTs have emerged to be a powerful prompting paradigm in the vision domain, such as objective detection~\cite{wu2024dettoolchain} and segmentation~\cite{li2024cpseg}. However, there is a lack of Chain-of-Thought methods tailored for OR tasks. Since an OR task can be treated as a coarse-to-fine problem~\cite{wang2023ord2seq}, our approach subtly designs a range optimization CoT, which is the first to propose using a CoT prompting paradigm to guide MLLMs for OR tasks.

\subsection{Ordinal Regression} 
Given an input image, ordinal regression in computer vision aims to map the input to a rank or a continuous value. Many popular methods~\cite{rothe2018deep,geng2013facial,frank2001simple,li2006ordinal,chen2017using} adopt a classification framework. Many recent studies~\cite{lim2019order, liu2019probabilistic, lee2020deep, li2021learning} proposed ordinal distribution constraints to exploit the ordinal nature of regression. Adding prior order knowledge to loss calculation, several methods~\cite{fu2018deep, diaz2019soft} created soft labels artificially by changing the distances between categories. A few advanced methods~\cite{liu2017deep, liu2018constrained, li2021learning, shin2022moving} sorted tuples that are formed by two or three instances with ordinal categories to learn the rank information. Ord2Seq~\cite{wang2023ord2seq} proposed to transform OR tasks as sequence prediction and solve ordinal regression using autoregressive models. Recent works like OrdinalCLIP~\cite{li2022ordinalclip}, L2RCLIP~\cite{wang2023learning}, and NumCLIP~\cite{du2024teach} used CLIP~\cite{CLIP} for OR tasks, focusing on designing the text encoder to map numerical labels to a continuous space for improved image-text alignment. Although these DL methods are general and effective, they need to train separate models for different OR tasks. This work explores utilizing MLLMs with our RO-CoT prompting paradigm to construct a unified OR model.

\section{Method}

\subsection{Overview}

Our work is the first to explore the effectiveness of MLLMs for OR tasks. Based on MLLMs, vision-language-driven OR tasks can be formulated as: Given a multimodal input (including images $I$, text $L$, etc.), output a classification result $r \in S$ within a candidate category set $S=\{C_1, C_2, \ldots,C_{n}\}$, where the candidate category labels are ordered, that is, $C_1 < C_2 < \cdots <C_n$. Based on our exploration that found the poor performance of MLLMs on OR tasks with training-free inference, we propose a novel OrderChain method to improve the ordinal understanding of MLLMs for effective ordinal regression. Fig.~\ref{fig2} shows an overview of OrderChain. OrderChian introduces a Range Optimization Chain-of-Thought (Sec.~\ref{roc}), a Category Recursive Division Method (Sec.~\ref{crdt}), and Task-aware prompts (Sec.~\ref{tap}) to conduct and reason the coarse-to-fine ordinal understanding process.

\subsection{Range Optimization Chain-of-Thought}
\label{roc}
Since OR tasks have continuous ordinal labels, a coarse-to-fine paradigm can be utilized to predict ordinal labels progressively. Inspired by this observation, we introduce a new prompting paradigm, called Range Optimization Chain-of-Thought (RO-CoT), to enable the MLLM to solve OR tasks in a progressive range optimization manner. Specifically, the procedure of our RO-CoT consists of four parts:

\begin{enumerate}[1.]
    \item {\bf Formatting and Setting Template}: Format individualized raw query inputs to a specific template of common OR tasks and set corresponding recursive CoT logic. 
    \item {\bf Definition Generation as Domain Knowledge Prompts}: For different OR tasks and user requirements, definitions and ideas of the tasks are generated through preliminary identification by LLM, which act as domain knowledge prompts to guide ensuing MLLM prediction.
    \item {\bf Loop Iteration in Each RO-CoT Step}: 
    \begin{enumerate}[(a)]
        \item {\bf Generating category feature prompts by the category recursive division (CRD) method} (discuss in Sec.~\ref{crdt}): Given the initial optional categories or last intermediate thought, CRD will continue to divide these categories into multiple consecutive subsets with a smaller range as new candidate categories for the next step. These candidate subsets will serve as category feature prompts for instruction. 
        \item {\bf Using combined task-aware prompts to instruct MLLM} (discuss in Sec.~\ref{tap}): The task-aware prompts, combined with default domain knowledge prompts and newly generated category feature prompts, are used to instruct the MLLM to propose a thought which a new candidate subset that the category of the image belongs to. 
        \item {\bf Continue or end}: If the final prediction is made, MLLM will return the outputs; otherwise, MLLM will repeat step (a) to continue making refined predictions with the new prompts generated by CRD. 
    \end{enumerate}
    \item {\bf Returning the Final Response}: When the subset contains only one category, the procedure will end and return the final prediction. 
    
\end{enumerate}

\begin{figure}[t]
\centering
\includegraphics[width=0.48\textwidth]{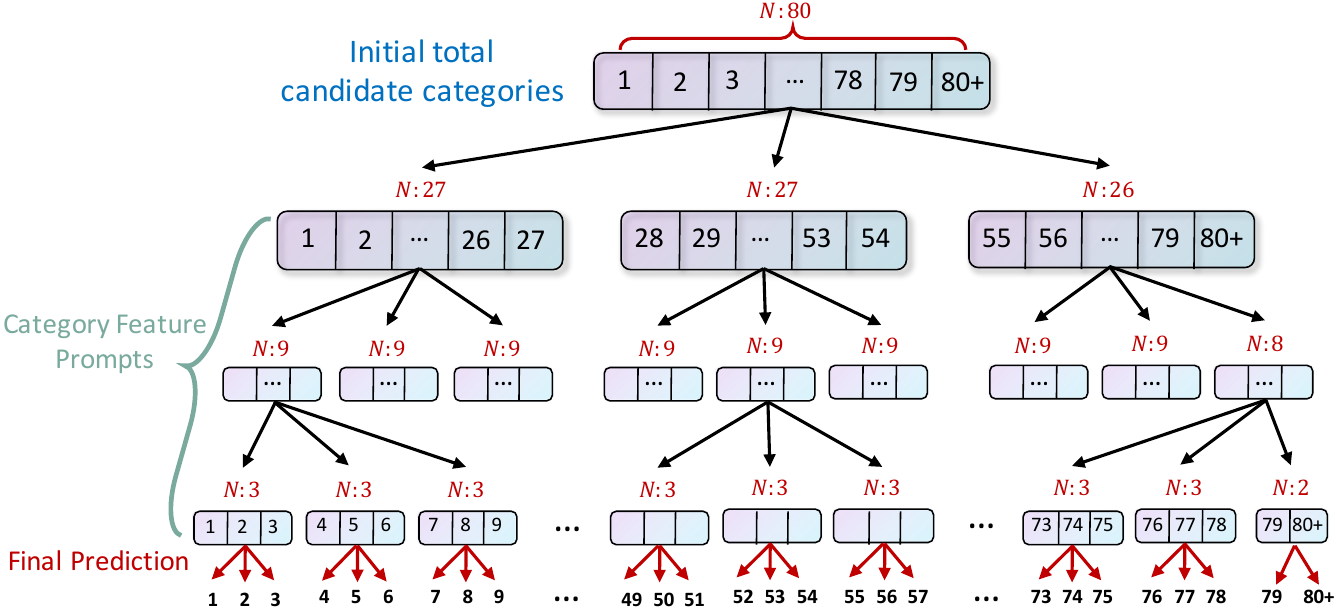}
\caption{An example of the Category Recursive Division (CRD) process on the facial age estimation task. Assume that the total number of categories is 80 and CRD uses a trinomial balanced tree, that is, each node has three child nodes (subsets) and the difference in the number of categories per child node (sub candidate sets) is minimal. CRD first divides the entire range (80) of categories into three subsets with new category numbers 27, 27 and 26. It repeats this process until the number of categories for each subset is less than or equal to 3. Upon the last division, the final prediction is obtained.  $N$ at each node denotes the number of categories within the node (candidate set).}
\label{figp}
\vskip -1 em
\end{figure}

\subsection{Category Recursive Division Method}
\label{crdt}

To achieve range optimization to enhance the understanding of category order, we introduce a Category Recursive Division (CRD) Method to automatically plan the path of range optimization.  CRD aims to divide the entire candidate categories into multiple more refined subsets. In this process, the range of candidate categories is optimized. Note that this strategy can be effective only when the categories are ordered, and is not suitable with non-order common classification tasks. Thus, this recursive division process can inject ordinal knowledge into the MLLM. Based on the automatic division, this method can limit as well as provide candidate options for each MLLM prediction. Specifically, assuming that the MLLM predicts a certain range, the CRD automatically queries the corresponding subsequent divisions, acting as a part of \texttt{Category Feature Prompts} to provide the MLLM with more refined candidate options and force the MLLM to focus on further refinement. \textcolor{black}{To avoid the negative impact of class unbalance, we structure the division process into a balanced division tree. Based on the initial total number of categories, $N_{init}$, we use a $k$-tree for division. The total recursive steps, $T$, is calculated as:
\vspace{-2pt}
\begin{equation}
    T = \log_{k}(N_{init}).
\end{equation}
\vskip -0.2 em
\noindent
For every step $i$, the maximum category number for each sub candidate set should be:
\vspace{-2pt}
\begin{equation}
    N_{i} = \frac{N_{init}}{k^{i}}+1, \quad  i =1,2,\ldots,T.
\end{equation}
\vskip -0.2 em
\noindent
Thus, the $j$-th candidate set $c_{i,j}$ of step $i$ should be:
\vspace{-2pt}
\begin{equation}
    c_{i,j} = \{s_{j}, s_{j}+1,\ldots, \min(N_{init}, s_{j}+N_{i}-1)\}, \\ 
\end{equation}
\vskip -0.2 em
\noindent
where the index of the starting category, $s_{j}$, for the candidate set  $c_{i,j}$ is:
\vspace{-2pt}
\begin{equation}
    s_{j} = (j-1) * N_{i}+1, \quad j =1,2,\ldots,k^i.
\end{equation}
\vskip -0.2 em
 An example is shown in Fig.~\ref{figp}.} In general, for datasets with fewer categories (e.g., image aesthetic assessment), we can use a binary tree.  For datasets with more categories (e.g., face age estimation), we tend to use a trinomial tree to reduce the CoT length and number of thoughts. This method can be applied to all OR tasks, helping develop a general ordinal understanding approach for MLLMs to learn and model the commonality of ordinal regression, that is, the intrinsic ordinal relations among the categories.

\begin{figure}[t]
\centering
\includegraphics[width=0.47\textwidth]{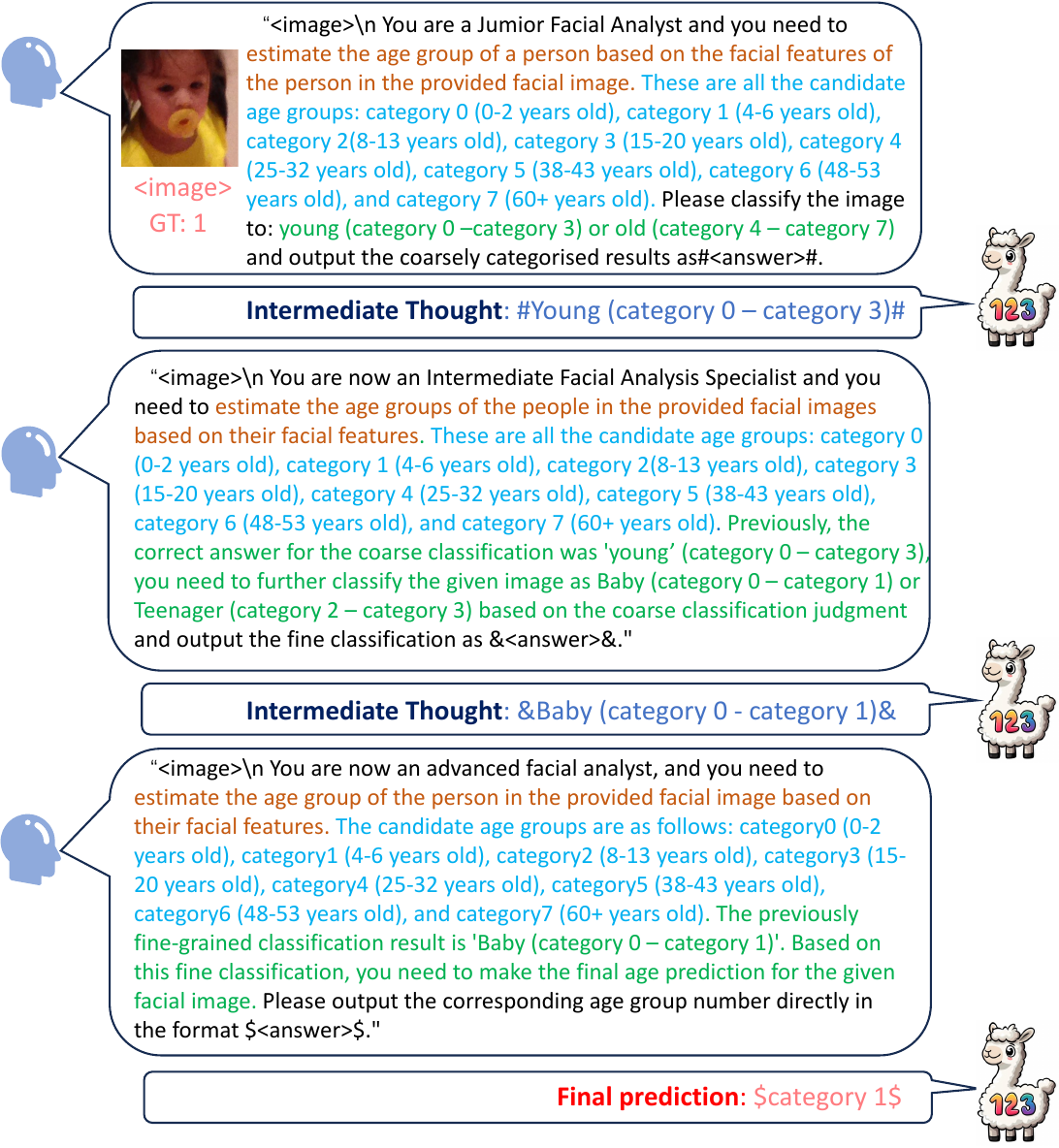}
\caption{An example of OrderChain for facial age estimation on the Adience dataset. \textcolor[RGB]{197,90,17}{Brown: Domain knowledge prompts}. \textcolor[RGB]{0,176,240}{Blue: Description prompts}. \textcolor[RGB]{0,176,80}{Green: Instruction prompts}. }
\label{fig4}
\end{figure}

\subsection{Task-aware Prompts}
\label{tap}

In the past, many general algorithms for ordinal regression have been proposed. Although effective in modeling common intrinsic order logic, they still required to train separate models for diverse OR tasks since the number and range of categories are objectively different. Subjectively, specific domain knowledge in different OR tasks is also difficult to model. Our observation is that the extensibility of MLLMs makes it possible to train a general OR model via prompt engineering. Based on this observation, we introduce task-aware prompts for OrderChain to enhance the modeling of the specificity of different OR tasks. Task-aware prompts contain two types: Category Feature Prompts and Domain Knowledge Prompts, which are elaborated on below.

\noindent
{\bf Category Feature Prompts.} To allow MLLMs to know the prediction objective and operate following the range optimization Chain-of-Thought, we introduce category feature prompts with two parts of prompts, focusing on different aspects: {\bf Description prompts} for describing the definition of the total categories and {\bf Instruction prompts} for candidate categories instruction. The details are given below.

\begin{enumerate}[1.]
    \item {\bf Description prompts} include the range and number of categories for a specific task (see the \textcolor{blue}{blue} part of Category Feature Prompts in Fig.~\ref{fig2}), for overcoming the disadvantage that the dimension of fully connected layers of the traditional DL models cannot be changed. By setting different description prompts, it is promising to train a universal MLLM to handle all OR problems.
    \item {\bf Instruction prompts} include candidate categories need to be refined at this step (see the \textcolor[RGB]{0,176,80}{green} part of Category Feature Prompts in Fig.~\ref{fig2}). \textcolor{black}{These prompts are obtained by the \texttt{Category Recursive Division} method based on previous predictions, and they act as the new query aiming to provide new limiting candidate categories for the MLLM to predict at this step. }
\end{enumerate} 

\begin{table}
  \centering
\scalebox{0.93}{
  \begin{tabular}{@{\quad}lcc@{\quad}}
    \toprule
    Method &  Accuracy (\%) $\uparrow$ & MAE $\downarrow$  \\
    \midrule
    \textit{Supervised SOTA}  & & \\
    \midrule
    CNNPOR~\cite{liu2018constrained}  &  57.4 & 0.55  \\
    GP-DNNOR~\cite{liu2019probabilistic}  & 57.4 & 0.54  \\
    SORD~\cite{diaz2019soft}  & 59.6 & 0.49  \\
    POE~\cite{li2021learning}  & 60.5 & 0.47 \\
    OrdinalCLIP~\cite{li2022ordinalclip}  & 61.2 & 0.47 \\
    MWR~\cite{shin2022moving}  & 62.6 & 0.45  \\
    Ord2Seq~\cite{wang2023ord2seq}  & 63.9 & 0.43 \\
    L2RCLIP~\cite{wang2023learning}  & 66.2 & 0.36 \\
    \midrule
    \textit{Zero-shot MLLM} & & \\
    \midrule
    LLaVA-1.5~\cite{liu2024improved}  & 17.6 & 1.48 \\  
    \midrule
    \textit{Lora Fine-tune MLLM~\cite{hu2022lora}}  & & \\
    \midrule 
    LLaVA-1.5 (baseline)~\cite{liu2024improved}  & 47.5 & 0.59  \\ 

    LLaVA-1.5 + OrderChain  & {\bf 93.2}   & {\bf 0.12}  \\

    \bottomrule
  \end{tabular}
  }
  
  \caption{Accuracy and MAE comparison on the Adience dataset.}
  \label{tab:a}
  \vskip -1 em
\end{table}

\noindent
{\bf Domain Knowledge Prompts.} Though the internal ordinal logic of different OR tasks is the same, the data features, estimation criteria, and so on can be very different. To model the specificity of different OR tasks, we introduce domain knowledge prompts, which are obtained by the MLLM itself and provide prior domain knowledge for the following predictions.  This is equivalent to a preliminary identification of the task, so that the MLLM can search for information related to the task as much as possible as a guide.

\section{Experiments}

\begin{table}
  \centering
 \scalebox{0.93}{
  \begin{tabular}{@{\quad}lcc@{\quad}}
    \toprule
    Method & Accuracy (\%) $\uparrow$
& MAE $\downarrow$ \\
    \midrule
    Palermo et al.~\cite{palermo2012dating} & 44.9 & 0.93 \\ 
    CNNPOR~\cite{liu2018constrained}  & 50.1 & 0.82  \\
    GP-DNNOR~\cite{liu2019probabilistic}  & 46.6 & 0.76 \\
    SORD~\cite{diaz2019soft}   & 53.4 & 0.70 \\
    POE~\cite{li2021learning}  & 54.7 & 0.66 \\
    MWR~\cite{shin2022moving}  & 57.8 & 0.58 \\
    Ord2Seq~\cite{wang2023ord2seq} & 60.9 & 0.52 \\
    OrdinalCLIP~\cite{li2022ordinalclip} & 56.4 & 0.67 \\
    L2RCLIP~\cite{wang2023learning} & 67.2 & 0.59 \\
    NumCLIP~\cite{du2024teach} & 69.6 & 0.35  \\
    \midrule

    LLaVA-1.5 (zero-shot)~\cite{liu2024improved} & 27.5 & 1.20 \\ 
    \midrule 
    
    LLaVA-1.5 (baseline)~\cite{liu2024improved} & 61.4 & 0.50 \\
    LLaVA-1.5 + OrderChain & {\bf 73.0} & {\bf 0.32} \\



    \bottomrule
  \end{tabular}
  }
  
  \caption{Accuracy and MAE comparison on the HCI dataset.}
  \label{tab:5}
 \vskip -1 em
\end{table}

\subsection{Datasets and Setup}

{\bf Datasets.}
To validate the effectiveness of our OrderChain, we conduct experiments on OR tasks in various domains, including Facial Age Estimation, Historical Image Dating, Image Aesthetics Assessment, and Diabetic Retinopathy Grading. The datasets for these tasks are as follows.

\begin{itemize}
    \item {\bf Facial Age Estimation.} We use the Adience dataset~\cite{levi2015age} for age group estimation that contains about 26,580 face images of 2,284 subjects from Flickr. Ages are annotated in 8 groups: 0-2, 4-6, 8-13, 15-20, 25-32, 38-43, 48-53, and over 60 years old. 

    \item {\bf Historical Image Dating.} The historical color image (HCI) dataset~\cite{palermo2012dating} is for estimating the decades of historical color photos. There are five decades from 1930s to 1970s, annotated as 1 to 5. Each decade has 265 images. 
    \item {\bf Image Aesthetics Assessment.} The Aesthetics dataset~\cite{dosovitskiy2020image} contains 15,687 Flickr image URLs, 13,706 of which are available. The dataset is used to grade image aesthetics. There are four image classes: animal, urban, people, and nature. Each image was graded by at least 5 different graders in 5 ranking categories to evaluate the photographic aesthetic quality: unacceptable, flawed, ordinary, professional, and exceptional. The ground truth is defined as the median rank among all the gradings.  

    \item {\bf Diabetic Retinopathy Grading.} The Diabetic Retinopathy (DR) dataset~\cite{cheng2023robust} contains 35,126 high-resolution fundus images. In this dataset, images were annotated in five levels of diabetic retinopathy from 1 to 5, representing no DR, mild DR, moderate DR, severe DR, and proliferative DR, respectively. 
\end{itemize}

\noindent
{\bf Experimental Setup.} Our experiments are conducted on the PyTorch platform with an NVIDIA Tesla A100 GPU. 
We use LLaVA-1.5-7B as our MLLM backbone, in which the image encoder is ViT-L-16 pre-trained by CLIP. We apply LoRA~\cite{hu2022lora} to fine-tune MLLM. We employ the AdamW~\cite{kingma2014adam} optimizer with a learning rate of 2e-4. We use a per-device batch size of 16. For fair comparison, all the known methods are implemented using their authors’ code or re-implemented based on the original papers.  The division of all the datasets follows~\cite{shin2022moving,wang2023ord2seq,li2022ordinalclip}.  More details of the datasets, experimental settings, and codes are given in the Supplemental Document.

\begin{table}
  \centering
\scalebox{0.95}{
  \begin{tabular}{@{\quad}lcc@{\quad}}
    \toprule
    Method & Accuracy (\%) $\uparrow$ & MAE $\downarrow$ \\
    \midrule
    Poisson~\cite{beckham2017unimodal} & 77.1 & 0.38 \\ 
    MT~\cite{ratner2018learning} & 82.8 & 0.36 \\

    SORD~\cite{diaz2019soft} & 78.2 & 0.73   \\
    POE~\cite{li2021learning} & 80.5 & 0.30  \\
    CIG~\cite{cheng2023robust} & 83.3 & 0.30 \\
    Ord2Seq~\cite{wang2023ord2seq} &  84.2 &  0.25 \\
    
    \midrule
    LLaVA-1.5 (zero-shot)~\cite{liu2024improved} & 3.9 & 12.1 \\ 
    \midrule 
    
    LLaVA-1.5 (baseline)~\cite{liu2024improved} & 30.0 & 0.99 \\
    LLaVA-1.5 + OrderChain & {\bf 85.7}& {\bf 0.23} \\

    \bottomrule
  \end{tabular}
  }
  
  \caption{Accuracy and MAE comparison on the DR dataset.}
  \label{tab:4}
  \vskip -1 em
\end{table}

\begin{table*}
  \centering
    \scalebox{0.93}{
  \begin{tabular*}{18.1cm}{lcccccccccc}
    \toprule
     \multirow{2}*{Method} &
    \multicolumn{5}{c}{Accuracy (\%) $\uparrow$ } 
    &  \multicolumn{5}{c}{MAE $\downarrow$}
    \\
    \cmidrule(r){2-6} \cmidrule(r){7-11}  & Nature & Animal & Urban & People & Overall 
    & Nature & Animal & Urban & People & Overall
    \\
    \midrule
    CNNPOR~\cite{liu2018constrained} & 71.86 & 69.32 & 69.09 & 69.94 & 70.05 & 0.294 & 0.322 & 0.325 & 0.321 & 0.316 \\
    SORD~\cite{diaz2019soft} & 73.59 & 70.29 & 73.25 & 70.59 & 72.03 & 0.271 & 0.308 & 0.276 & 0.309 & 0.290 \\
    POE~\cite{li2021learning}  & 73.62 & 71.14 & 72.78 &  72.22 & 72.44 & 0.273 & 0.299 & 0.281 &  0.293 & 0.287\\
    Ord2Seq~\cite{wang2023ord2seq} & {\bf 78.09}	& {\bf 75.74} & 72.83 & 69.24 & \textcolor{cyan}{74.43} & {\bf 0.225} & \textcolor{cyan}{0.257} & 0.275 & 0.319 &  0.264 \\ 
    OrdinalCLIP~\cite{li2022ordinalclip}  & 73.65 & 72.85 & 73.20  & \textcolor{cyan}{72.50} &  73.05 & 0.273 & 0.279 &0.277 & \textcolor{cyan}{0.291} & 0.280\\
    L2RCLIP~\cite{wang2023learning}  & 73.51 & \textcolor{cyan}{75.26} & \textcolor{cyan}{77.76} & {\bf 78.69} & \textcolor{orange}{76.07} & 0.267&  \textcolor{orange}{0.253} & \textcolor{orange}{0.216} & \textcolor{orange}{0.246} & \textcolor{orange}{0.245}\\
    NumCLIP~\cite{du2024teach}  & \textcolor{orange}{75.20} & 75.24 & {\bf 79.49} & \textcolor{orange}{76.17} & {\bf 76.53} & \textcolor{cyan}{0.249} & {\bf 0.250} & {\bf 0.208} & {\bf 0.238} & {\bf 0.236}\\
    
    \midrule
    LLaVA-1.5 (zero-shot)~\cite{liu2024improved} & 3.68 & 8.01 & 10.8 & 1.47 &  5.21 & 1.422 & 1.109 & 1.439 & 0.901 & 1.275 \\
    \midrule 
    LLaVA-1.5 (baseline)~\cite{liu2024improved} & 41.81 & 64.86 & 61.89 & 39.33 & 59.56 & 0.418 & 0.374 & 0.393 & 0.597 & 0.435 \\

     LLaVA-1.5 + OrderChain & \textcolor{cyan}{73.91} & \textcolor{orange}{75.61} & \textcolor{cyan}{77.87} & 66.26 & 73.83 & \textcolor{orange}{0.229} & 0.260 & \textcolor{cyan}{0.252} & 0.297 & {\textcolor{cyan}{0.260}} \\

    \bottomrule
  \end{tabular*}
  }
  
  \caption{Results on the Image Aesthetics dataset. Accuracy and MAE are reported for each of the four image classes. The best three results are marked in {\bf bold}, \textcolor{orange}{orange}, and \textcolor{cyan}{blue}, respectively.}
  \label{tab:3}
  \vskip -1 em
\end{table*}

\subsection{Experimental Results}

\noindent
{\bf Facial Age Estimation.} 
Fig.~\ref{fig4} gives an example of OrderChain for facial age estimation on the Adience dataset, and Table~\ref{tab:a} shows comparison results with state-of-the-art (SOTA) methods on the Adience dataset.  One can see that the zero-shot version (training-free) and baseline version (LoRA fine-tuning) of LLaVA-1.5 perform worse than known supervised SOTA methods. Specifically, the zero-shot version of LLaVA-1.5 attains only 17.6\% accuracy and baseline version of LLaVA -1.5 perform better but merely obtains a mediocre performance. In contrast, LLaVA with our proposed OrderChain achieves 93.2\% accuracy and 0.12 MAE, outperforming SOTA methods L2RCLIP~\cite{wang2023learning} by a remarkable margin of $\sim$27\% improvement in accuracy and 0.24 reduction in MAE.  This indicates that when integrated with our OrderChain, the MLLM is instructed to think of the facial age estimation problem step by step with a few smaller refined subtasks, which allows the MLLM to learn the internal ordinal relationships between categories and effectively estimate the ages of faces with a wide range and a large number of categories. Our Ordchain achieves milestone performance on age estimation tasks, which further demonstrates the effectiveness of OrderChain in improving the ordinal understanding of MLLM. 

\noindent
{\bf Historical Image Dating.} Table~\ref{tab:5} compares the results on the HCI dataset. As can be seen, LLaVA with our OrderChain outperforms known methods and achieves state-of-the-art results, yielding improvements of 3.4\% in Accuracy and 0.03 in MAE, which indicate the superiority of our new approach. Compared to the zero-shot and baseline versions of LLaVA, LLaVA with our OrderChain achieves superior performance with remarkable improvement, validating the limitation of vanilla LLaVA and that the main improvement comes from our proposed OrderChain. In addition, we find that, like the facial age estimation task, the labels of the HCI dataset are also objective and true. For this kind of OR task, the MLLM has great potential and gains large improvement with our proposed OrderChain, indicating that OrderChain can learn a conceptual understanding of the essential order of objects in OR tasks.

\noindent
{\bf Diabetic Retinopathy Grading.}
Table~\ref{tab:4} shows the results on the DR dataset. Note that this dataset is unbalanced since the sample number decreases sharply as the severity DR level increases. Known methods yield poor performances due to the unbalanced data, like SORD~\cite{diaz2019soft}, which is a modality-specific method utilizing modified soft labels, suffers serious errors in MAE. Worse, the zero-shot and baseline versions of LLaVA attain horrible performance. In comparison, LLaVA with our OrderChain still maintains competitive performances, achieving an Accuracy of 85.7\% and an MAE of 0.23, which greatly outperforms the baselines and the other order learning methods, showing that our approach has better robustness on unbalanced data. It is mainly due to the category range division process of RO-CoT, which also enables better positive-negative distinction. Unlike one positive class against the other negative classes in previous work, it turns to classifying the first two categories against the last three categories in the first CoT step of OrderChain for the DR dataset (5 categories in total). In this way, the classification in a step is more category-balanced, which helps to better deal with unbalanced data. Moreover, our OrderChain provides a promising way to grade medical diseases of the OR type by a unified MLLM.

\begin{figure}[t]
\centering
\includegraphics[width=0.48\textwidth]{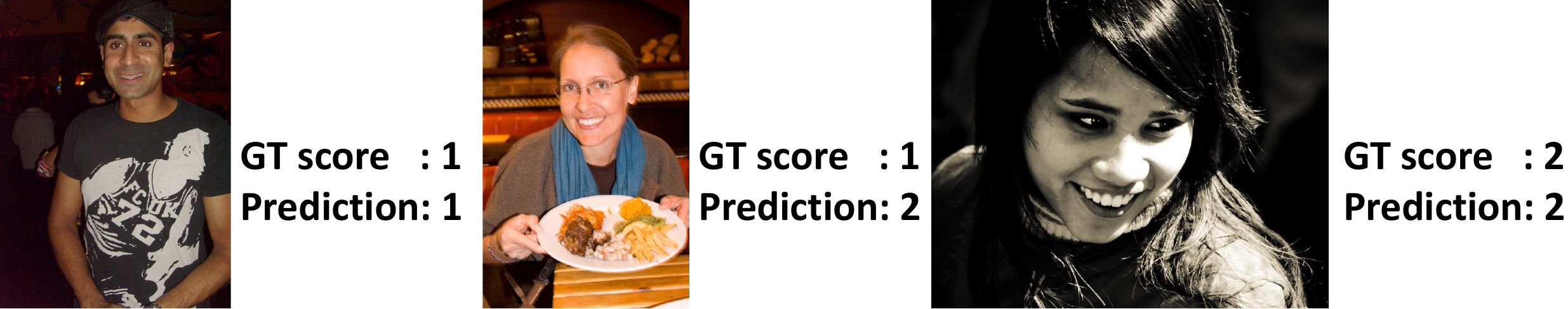}
\caption{Quantitative visualization results of some examples of people aesthetics assessment.}
\label{fig:re1}
\vskip -1 em
\end{figure}

\noindent
{\bf Image Aesthetics Assessment.} Table~\ref{tab:3} shows the results on the Image Aesthetics dataset. \textbf{We find that the MLLM with our OrderChain does not achieve optimal performance, which we believe is due to the highly {\it subjective} nature of the labels made by human raters. The subjective differences between different people, and even between people and MLLMs for the definition of beauty, may be significant. Especially for the relatively low performance on the `People' category, we suspect it is due to the pre-training imposed on the MLLM that tends to praise rather than demean people. To further illustrate this point, we provide quantitative visualization results as shown in the F sig.~\ref{fig:re1}. Our model successfully predicted the aesthetic scores for the first and third images. However, for the second image, the model makes an incorrect prediction with a higher score, which also reflects the issue we described in the paper: MLLMS have positive biases towards people.}. In other relatively more objective categories, LLaVA with our OrderChain achieves higher performance, and thus the overall performance is mainly affected by the `People' category. On the other hand, our proposed OrderChain remarkably improves vanilla LLaVA to a competitive level, demonstrating the effectiveness of our OrderChain.

\begin{table*}[t]
\centering
\scalebox{0.92}{
\begin{tabular}{lll}
\toprule
Method & Accuracy (\%) $\uparrow$ & MAE $\downarrow$  \\ \midrule
(a) LLaVA (zero-shot)      & \quad 17.6          & \ 1.48\\
\midrule
(b) LLaVA (baseline)   & \quad 47.5\quad       & \ 0.59 \quad \\ 
(c) LLaVA + \texttt{Domain Knowledge Prompts}  &  \quad  58.0\quad(\textcolor{dg}{+10.5})  & \ 0.49 \quad (\textcolor{dg}{-0.10})  \\ 
(d) LLaVA +  \texttt{Category Feature Prompts} & \quad {32.5}\quad(\textcolor{red}{-15.0})     & \ {1.42} \quad (\textcolor{red}{+0.83})\\ 
(e) LLaVA +  \texttt{Domain Knowledge Prompts} + \texttt{Category Feature Prompts} & \quad {38.7}\quad(\textcolor{red}{-8.8})     & \ {1.35} \quad (\textcolor{red}{+0.76})\\  
(f) LLaVA + \texttt{Category Feature Prompts} + \texttt{RO-CoT} &\quad {84.6}\quad(\textcolor{dg}{+37.1})     & \ {0.18} \quad (\textcolor{dg}{-0.41})\\ 
\midrule
(g) LLaVA + OrderChain  & \quad {\bf 93.2}\quad(\textcolor{dg}{+45.7})     & \ {\bf 0.12} \quad (\textcolor{dg}{-0.47})\\ 
\bottomrule
\end{tabular}%
}
\caption{Ablation experiments on the Adience dataset. 
RO-CoT denotes range optimization Chain-of-Thought. }
\label{tab:11}
\vskip -1 em
\end{table*}

\subsection{Ablation Study}
We conduct a comprehensive ablation study to examine the effectiveness of each key component in our OrderChain, including Domain Knowledge Prompts, Category Feature Prompts, and Range Optimization Chain-of-Thought. Except for the zero-shot version, all the experiments presented in this subsection are conducted on the Adience dataset and use LoRA to fine-tune.   Specifically, LLaVA (zero-shot) denotes directly using LLaVA for training-free inference. LLaVa (baseline) denotes using LLaVA for fine-tuning based on standard image-label pair samples.  Table~\ref{tab:11} shows the results, from which several observations can be drawn. {\bf (1)} LLaVA (zero-shot) attains very limited performance, which indicates that the MLLM is difficult to exert the ability of ordinal understanding in the training-free situation. {\bf (2)} LLaVA (baseline), which is fine-tuned by image-label pairs, achieves normal performance but still leaves potential for improvement to be desired. {\bf (3)} Compared to the baseline LLaVA, the addition of \texttt{Domain Knowledge Prompts} provides considerable performance gains of nearly 10\% in accuracy, demonstrating the importance of task-specific modeling. {\bf (4)} By comparing (b) and (d) in Table~\ref{tab:11}, as well as (c) and (f), we find that merely adding \texttt{Category Feature Prompts} without RO-CoT could not help the MLLM to improve, which we hypothesize is due to the lack of relationships of multi-stage prompts that confuses the MLLM in the absence of CoT. {\bf (5)} Comparing (f) and (d), it shows the remarkable improvement of the MLLM brought about by RO-CoT based on \texttt{Category Feature Prompts}, which demonstrates that our RO-CoT can fully utilize these prompts for ordinal understanding, since RO-CoT can rigorously connect multi-stage refined prompts to build a common thinking paradigm for OR tasks. The full version (g), LLaVA with our OrderChain, achieves the best performance, proving our OrderChain's effectiveness in commonality and specificity modeling for endowing more powerful ordinal understanding to the MLLM.

\subsection{Zero-shot Generalization}
To further validate that our OrderChain learned the general way of thinking for ordinal understanding, we conduct inference-only experiments to show the zero-shot performance of baseline LLaVA and our OrderChain, as shown in Tab.~\ref{tab:n1}. We choose three additional datasets including two in-domain datasets (APTOS~\cite{karthik2019aptos} dataset for DR grading and AVA dataset for aesthetic assessment~\cite{gu2018ava}) 
and one out-of-domain dataset (KonIQ~\cite{hosu2020koniq} for image quality assessment). As we can see, our OrderChain can bring a remarkable zero-shot improvement to the baseline LLaVA on both in-domain and out-of-domain datasets, substantiating the generalization ability of OrderChain. Specifically, the significant improvement on the APTOS dataset can be attributed to to the model's prior training on similar tasks, enabling it to better adapt to the understanding of this specific task. However, even for unseen scenarios ({\it i.e.} image quality assessment), our OrderChain can still obtain a 15\% improvement compared to baseline LLaVA. This demonstrates that our approach indeed stimulates the ordinal understanding ability of MLLMs, equipping them with a generalized way of thinking for ordinal regression tasks.

\begin{table}[t]
\centering
\scalebox{0.82}{
\begin{tabular}{@{\quad}lcccc@{\quad}}
\toprule
Method & Metric &   APTOS~\cite{karthik2019aptos} & AVA~\cite{gu2018ava} & KonIQ~\cite{hosu2020koniq}  \\
\midrule
\multirow{2}{*}{Baseline} & ACC $\uparrow$ & 23.2 & 49.6 & 36.3\\
 & MAE $\downarrow$ & 1.02 & 0.56 & 0.81\\
\midrule
\multirow{2}{*}{OrderChain} & ACC $\uparrow$  & 70.0 & 57.7 & 51.5 \\ 
 & MAE $\downarrow$  & 0.30 & 0.44 & 0.49 \\ 
\bottomrule
\end{tabular}
}
\caption{Zero-shot Performance of baseline LLaVA and our Orderchain on in-domain and out-of-domain datasets. }
\label{tab:n1}
\vskip -1 em
\end{table}

\section{Conclusions}
In this paper, we presented a novel and general prompting paradigm, OrderChain, to improve the ordinal understanding of MLLMs for ordinal regression. We first pointed out two major reasons for the limited performance of vanilla
MLLMs, i.e., lack of specificity modeling and commonality modeling. We adopted a range optimization Chain-of-Thought to learn a commonality way of thinking about ordinal regression tasks and task-aware prompts to inject task-specific information into MLLMs. We also introduce a category recursive division tool to generate refined candidate category subsets for supporting range optimization. \textcolor{black}{Extensive experiments showed that our OrderChain significantly improves the performance of the MLLM and achieves optimal performance in most ordinal regression tasks and considerable zero-shot generalization performance, demonstrating that OrderChain can effectively improve the ordinal understanding of MLLMs and provides a promising paradigm to build a unified ordinal regression MLLM.}

\paragraph{Acknowledgements} This research was partially supported by National Natural Science Foundation of China under grants No. 82202984, No. 12326612, Zhejiang Key R\&D Program of China under grant No. 2024SSYS0026, and the Transvascular lmplantation Devices Research Institute (TIDRI).

{
    \small
    \bibliographystyle{ieeenat_fullname}
    \bibliography{main}
}

\end{document}